\documentclass[final]{cvpr}

\usepackage{times}
\usepackage{epsfig}
\usepackage{graphicx}
\usepackage{amsmath, amsbsy}
\usepackage{bm}
\usepackage{amssymb}
\usepackage{xcolor,colortbl}
\usepackage{subfigure}
\usepackage{pifont}
\usepackage{array}
\usepackage{bbold}
\usepackage{nicefrac}
\makeatletter
\@namedef{ver@everyshi.sty}{}
\makeatother
\usepackage{tikz}
\usepackage{tikzscale}
\usepackage{pgfplots}
\pgfplotsset{compat=newest}
\usepackage[mode=buildnew]{standalone}
\usepackage{booktabs}
\usepackage{multirow}
\usepackage{makecell}
\usepackage{siunitx}

\usepackage[pagebackref=true,breaklinks=true,colorlinks,bookmarks=false]{hyperref}

\def\cvprPaperID{1} 
\def\confYear{CVPR 2021}

\newdimen\owntablesep
\newcommand{\stz}{\rule{0mm}{2.3ex}}

\DeclareMathOperator*{\argmax}{argmax} 

\DeclareSymbolFont{AMSb}{U}{msb}{m}{n}
\DeclareSymbolFontAlphabet{\mathbb}{AMSb}
\begin{document}

\title{Improving Online Performance Prediction for Semantic Segmentation}

\author{Marvin Klingner\quad Andreas Bär\quad Marcel Mross\quad Tim Fingscheidt\\
{\tt\small \{m.klingner, andreas.baer, m.mross, t.fingscheidt\}@tu-bs.de}\\[1.0em]
Technische Universität Braunschweig, Braunschweig, Germany}

\maketitle

\begin{abstract}
    In this work we address the task of observing the performance of a semantic segmentation deep neural network (DNN) during online operation, i.e., during inference, which is of high importance in safety-critical applications such as autonomous driving. Here, many high-level decisions rely on such DNNs, which are usually evaluated offline, while their performance in online operation remains unknown. To solve this problem, we propose an improved online performance prediction scheme, building on a recently proposed concept of predicting the primary semantic segmentation task's performance. This can be achieved by evaluating the auxiliary task of monocular depth estimation with a measurement supplied by a LiDAR sensor and a subsequent regression to the semantic segmentation performance. In particular, we propose (i) sequential training methods for both tasks in a multi-task training setup, (ii) to share the encoder as well as parts of the decoder between both task's networks for improved efficiency, and (iii) a temporal statistics aggregation method, which significantly reduces the performance prediction error at the cost of a small algorithmic latency. Evaluation on the KITTI dataset shows that all three aspects improve the performance prediction compared to previous approaches.
\end{abstract}
 
\section{Introduction}
\label{sec:introduction}

\begin{figure}
    \centering
    \includegraphics[width=1.0\linewidth]{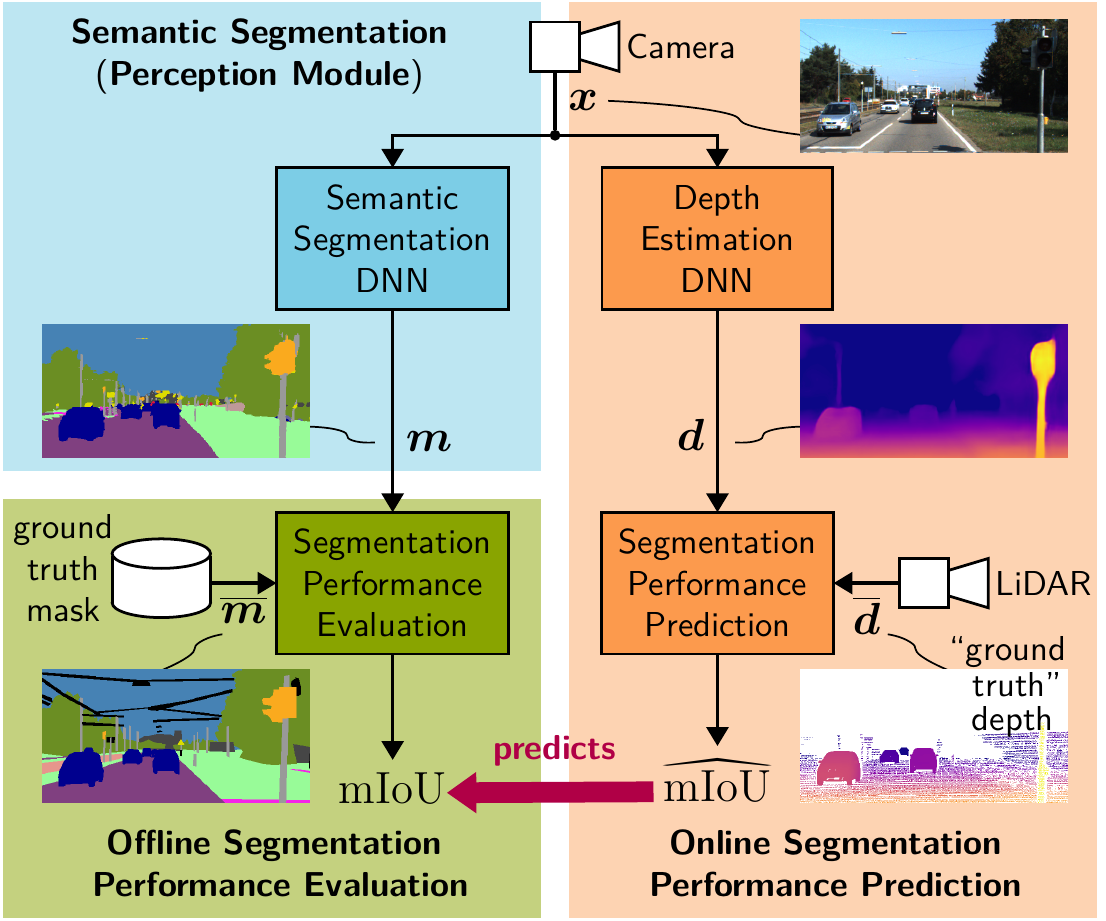}
    \caption{\textbf{Online performance prediction} ($\widehat{\mathrm{mIoU}}$) concept for the semantic segmentation task \textbf{during online inference}. The typical offline evaluation requiring ground truth data is replaced by the auxiliary depth estimation task, which can be evaluated online using a LiDAR measurement.}
    \label{fig:inference_concept}
\end{figure}

The task of semantic segmentation is an essential component for camera-based environment perception applied in safety-critical applications such as autonomous driving, remote sensing, or virtual reality. Here, high-level decision systems require reliable information from the components of the environment perception. While deep neural networks (DNNs) nowadays achieve state-of-the-art performance on the semantic segmentation task \cite{Chen2018a, Wang2019b, Zhu2019}, due to the lack of ground truth labels their performance during online deployment cannot be measured in a straightforward fashion. Moreover, the performance can drop significantly due to environment changes or domain shifts such as lighting changes, camera noise, or even adversarial attacks \cite{Carlini2017a, Goodfellow2015, Hussain2019, Madry2018, Zhou2019b}. Predicting the actual performance of semantic segmentation DNNs is therefore of highest interest.
\par
Semantic segmentation DNNs are usually evaluated offline on well-prepared datasets \cite{Cordts2016, Neuhold2017}. The observed offline performance (measured,\eg, by the $\mathrm{mIoU}$ a metric, cf.~Fig.~\ref{fig:inference_concept}) is assumed to be maintained in online operation, as a direct evaluation is not possible without labels. Some initial works intend to mitigate the problem by temporal or network ensemble consistency checks  \cite{Gal2016, Ritter2018, Varghese2020}, however, these approaches are not straightforward extendable to arbitrary tasks. Another approach uses an autoencoder to predict the performance based on the image reconstruction error \cite{Loehdefink2020}. This has, however, not been shown to be applicable on a few-shot basis, which would be necessary for online operation. Conclusively, there is still a desire for methods predicting DNN performance in online operation.
\par
Addressing this task of online performance prediction, we build on the work of Klingner~\etal~\cite{Klingner2021}, proposing a method, which relies on the auxiliary task of depth estimation also predicted by a DNN. As shown in Fig.~\ref{fig:inference_concept}, the semantic segmentation DNN (blue parts) is usually evaluated offline (green parts) by the mean intersection over union ($\mathrm{mIoU}$) metric. In contrast, for online performance prediction (orange parts), the same input image is also given to a depth estimation DNN. The resulting depth estimate is evaluated afterwards. Finally, the depth performance can be connected to the segmentation performance by a simple polynomial regression, giving a prediction for the actual segmentation performance.
\par
Our contribution with this work is threefold. Firstly, we introduce a \textit{sequential} training protocol for both tasks in a multi-task network, improving the absolute performance as well as the performance prediction. Secondly, we analyze how many layers should be shared between both tasks to increase their correlation and thereby the performance prediction while still retaining a high performance of the semantic segmentation. Sharing more layers in the decoder also reduces the \textit{computational complexity} induced by the performance prediction during inference, which can be of interest for computationally restricted settings. Thirdly, we investigate how a \textit{temporal aggregation of statistics} can improve the prediction performance at the cost of a small decision latency but without affecting the absolute segmentation performance. Our analysis on the KITTI dataset \cite{Geiger2013} will show that all three methods significantly reduce the performance prediction error.

\section{Related Work}
\label{sec:related_work}

In this section we give an overview on semantic segmentation techniques and how they profit from multi-task learning setups with depth estimation. Finally, we discuss other techniques for online performance prediction of DNNs.
\par 
\textbf{Semantic Segmentation}:
The task of semantic segmentation has been facilitated by the increasing availability of public datasets \cite{Cordts2016,Neuhold2017,Yu2018b} and the use of fully convolutional networks as outlined in the initial work of Long~\etal~\cite{Long2015}. Subsequent works adapted this concept and introduced improvements such as dilated convolutions \cite{Chen2015, Yu2016}, atrous spatial pyramid pooling \cite{Chen2018a}, state-of-the-art feature extractors \cite{Chen2018, Orsic2019}, label relaxation \cite{Zhu2019}, or high resolution networks \cite{Wang2019b}. Other approaches explore unsupervised domain adaptation techniques \cite{Klingner2020c, Kim2020, Wang2020, Yang2020a} to mitigate the domain shift problem for semantic segmentation and thereby facilitate the use of readily available synthetic datasets \cite{Richter2016,Ros2016}. Moreover, computational complexity can be reduced tremendously with only marginal loss in performance due to the use of efficient architectures. Examples are factorized convolutions \cite{Mehta2018, Romera2018} or depthwise-separable convolutions \cite{Chollet2017}, or a more efficient architectural design in general \cite{Li2019a, Orsic2019}. Another current research stream focuses on robustness of semantic segmentation DNNs to domain shifts or adversarial attacks \cite{Baer2020,Baer2021,Chen2019e,Hendrycks2019a,Liu2019c}. In this work we do not approach semantic segmentation from a pure performance point of view but introduce a concept for an efficient online performance prediction.
\par
\textbf{Multi-Task Learning}:
The joint learning of depth estimation and semantic segmentation, either by shared network structures \cite{Eigen2015,Kendall2018} or by combined loss functions \cite{Kendall2018, Xu2018c, Zhang2019a}, has been shown to be beneficial for both tasks. Here, initial multi-task learning settings made use of supervised depth estimation. However, recent focus of depth estimation research shifted to self-supervised depth estimation techniques \cite{Zhou2017a,Godard2017,Godard2019} due to the wide applicability of such methods not relying on ground truth depth labels. Conclusively, such techniques are also successfully combined with semantic segmentation \cite{Chen2019a,Guizilini2020,Novosel2019}. On the one hand, the segmentation can be used to reduce the influence of dynamic objects on the depth estimation model assuming a static world \cite{Klingner2020a,RaviKumar2021,RaviKumar2021a} and sharpen depth edges at object boundaries \cite{Chen2019a,Zhu2020}. On the other hand, the depth estimation gives additional structural information to the semantic segmentation model \cite{Chen2019a,Novosel2019} and this multi-task setup can even be beneficial for unsupervised domain adaptation \cite{Klingner2020a,Lee2019d,Vu2019}. Finally, cross-task consistency checks \cite{Yang2018b,Meng2019a} and pixel-adaptive convolutions can be used to improve the performance of both tasks \cite{Guizilini2020, RaviKumar2021}.
\par
While most works focus on performance gains of multi-task learning techniques, a recent work \cite{Klingner2020} also showed a positive effect on the robustness of a semantic segmentation DNN. Similarly, in this work we combine semantic segmentation with self-supervised depth estimation and show how this setting can be used to predict the performance of a semantic segmentation DNN in online operation.
\par
\textbf{Performance Prediction of DNNs}:
The usual way to evaluate the performance of a semantic segmentation DNN offline is to use a dataset containing ground truth labels \cite{Everingham2015, Cordts2016, lin2014microsoft}, which is not possible in an online setting. To mitigate this problem one can model uncertainty in the output, \eg, by Bayesian DNN approaches \cite{Gal2016, Blundell2015, Ritter2018} which, however, involves the training and inference of an ensemble of networks and is therefore computationally expensive. Alternatively, temporal consistency of consecutive predictions can be checked \cite{Varghese2020}, though the definition of consistency is highly task-specific. A recent successful approach of Löhdefink~\etal~\cite{Loehdefink2020} estimates the performance of a semantic segmentation model based on the image reconstruction error of an autoencoder DNN. Similarly, Klingner~\etal~\cite{Klingner2021} use the depth estimation task for this purpose. Both approaches, however, still have a high error in their segmentation performance predictions, which we significantly reduce in this work. In contrast to \cite{Loehdefink2020, Klingner2021} we therefore investigate, which training technique and multi-task network architecture is best suited for performance prediction. Also, we show how and under which conditions it is useful to employ a temporal aggregation of performance predictions.

\section{Method Description}
\label{sec:method_description}

Here, we first describe our multi-task training procedure for semantic segmentation and depth estimation. Afterwards, we describe our online performance prediction, which can be applied after both tasks have been trained. 

\subsection{DNN Training}

\begin{figure}
    \centering
    \includegraphics[width=1.0\linewidth]{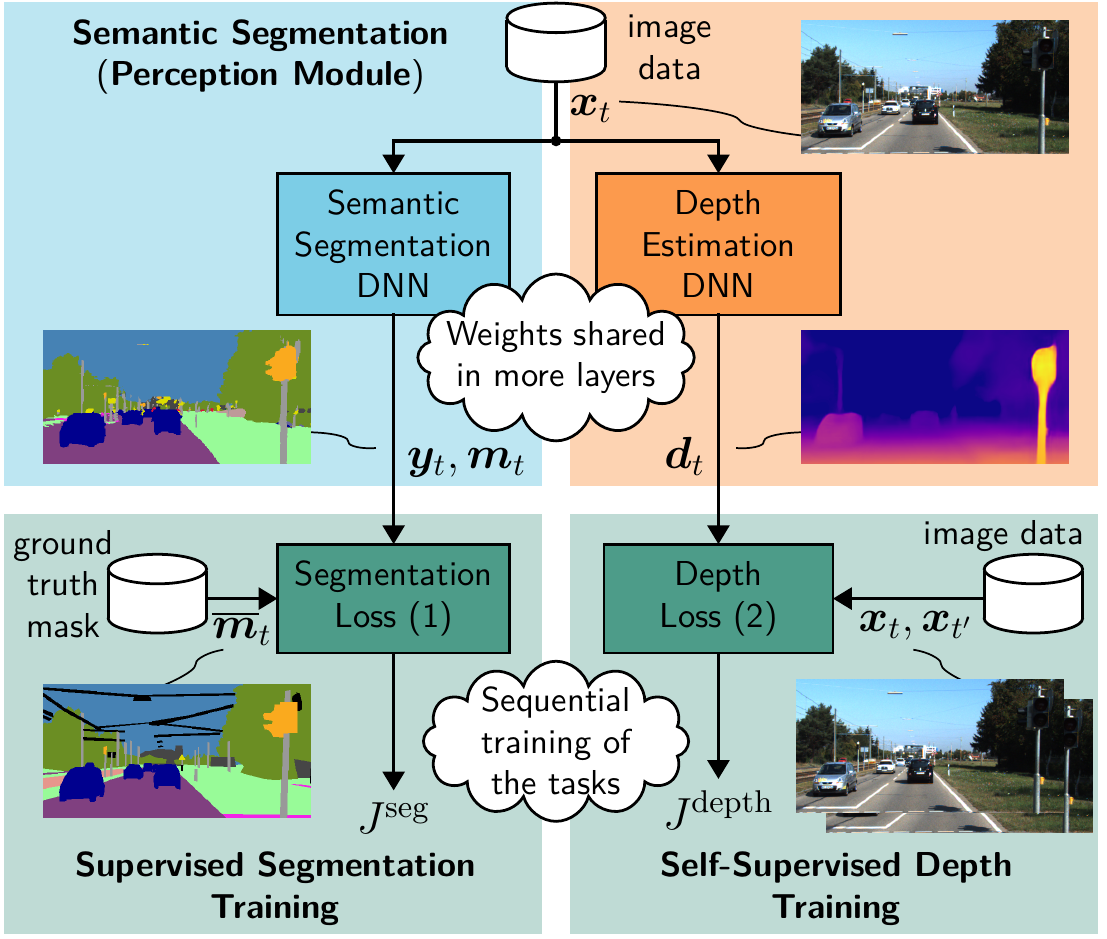}
    \caption{\textbf{Training setup} of the \textbf{online performance prediction} for the semantic segmentation task. Segmentation and depth tasks are initially trained by two separate loss functions. The final performance prediction can, however, be improved by sharing more network layers and---if a multi-task network is used---by training both tasks in a sequential fashion.}
    \label{fig:training_concept}
\end{figure}

\begin{figure}
    \centering
    \includegraphics[width=1.0\linewidth]{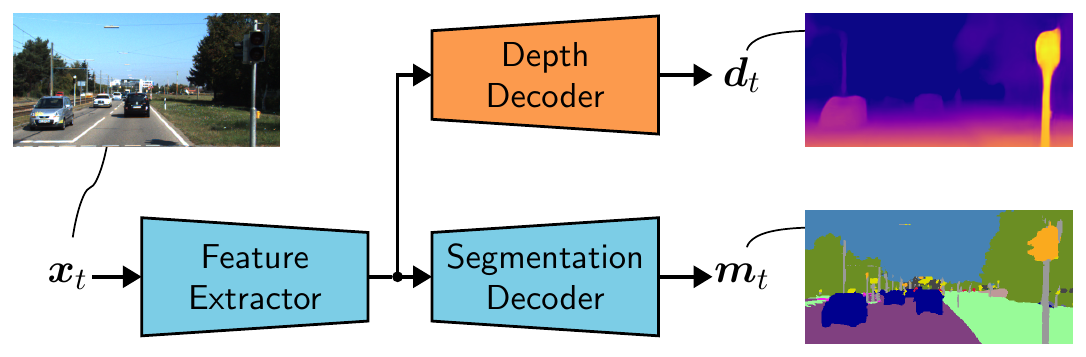}
    \caption{\textbf{Multi-task network architecture} used for the joint estimation of depth and segmentation outputs.}
    \label{fig:multi_task_network}
\end{figure}

As shown in Fig.~\ref{fig:training_concept}, we train both tasks simultaneously using task-specific loss functions. To find the optimal influence of each task, we do not weigh the loss functions, but the scaled gradients are flowing into the shared network parts (cf.~the multi-task network in Fig.~\ref{fig:multi_task_network}) as in \cite{Klingner2021, Klingner2020a}. 

\textbf{Semantic Segmentation}:
The semantic segmentation relies on a single input image $\bm{x}_t\in\mathbb{I}^{H\times W\times C}$, where $H$ defines the height, $W$ the width, and $C=3$ the number of color channels. The range $\mathbb{I} = \left[0,1\right]$ defines the normalized range of gray values. At inference, the neural network is used to predict a pixel-wise semantic segmentation map $\bm{m}_t \in \mathcal{S}^{H\times W}$ from the single image $\bm{x}_t$. Here, $\mathcal{S}$ is the set of considered semantic classes. The DNN is trained by supervision from ground truth labels $\overline{\bm{m}}_t= \left(\overline{m}_{t,i}\right)\in\mathcal{S}^{H\times W}$, which are utilized in a one-hot encoded fashion as $\overline{\bm{y}}_t = \left(\overline{y}_{t,i,s}\right)\in \left\lbrace 0,1\right\rbrace^{H\times\ W\times |\mathcal{S}|}$, such that $\overline{m}_{t,i} = \argmax_{s \in \mathcal{S}} \overline{y}_{t,i, s}$ at each pixel index $i \in \mathcal{I} = \left\lbrace 1, ..., H\cdot W\right\rbrace$ in the image. During training, the network then predicts output probabilities $\bm{y}_t = \left(y_{t,i, s}\right) \in\mathbb{I}^{H\times W\times |\mathcal{S}|}$ with $|\mathcal{S}|$ semantic classes, which are optimized utilizing the cross-entropy loss
\begin{equation}
J^{\mathrm{seg}} = -\frac{1}{H\cdot W}\sum_{i \in\mathcal{I}}\sum_{s \in\mathcal{S}} w_s \overline{y}_{t,i,s} \cdot \log\left(y_{t,i,s}\right), 
\label{eq:crossentropy_loss}
\end{equation}
where $w_s$ are the class weights as defined in \cite{Paszke2016}. The final pixel-wise classes are then obtained by $m_{t,i} = \argmax_{s \in \mathcal{S}} y_{t, i, s}$, yielding the estimated segmentation mask $\bm{m}_t = \left(m_{t,i}\right)\in \mathcal{S}^{H\times W}$.
\par
\textbf{Monocular Depth Estimation}: 
The task of monocular depth estimation is to assign a depth value $d_{t,i}\in\mathbb{D}$ to each input image pixel $\bm{x}_{t,i}\in \mathbb{I}^C$. Thereby, a depth map $\bm{d}_t \in \mathbb{D}^{H\times W}$ is predicted from a monocular input image $\bm{x}_t$. Here, $\mathbb{D} = \left[d_{\mathrm{min}},d_{\mathrm{max}}\right]$ defines the range of possible depth values, constrained by $d_{\mathrm{max}}$ and $d_{\mathrm{min}}$, representing an upper bound and a lower bound, respectively.
\par
To ensure that the depth estimation is applicable to a wide range of datasets, we build on recent works \cite{Zhou2017a, Godard2019}, who train the depth estimation in a self-supervised fashion on sequential video images. To this end we consider consecutive pairs of images ($\bm{x}_t$, $\bm{x}_{t'}$) with $t'\in\mathcal{T}' = \left\lbrace {t\!-\!1}, {t\!+\!1} \right\rbrace$, which are given to a second network, predicting the six degrees of freedom of the relative pose between both images $\bm{T}_{t\rightarrow t'}\in \mathit{SE}(3)$. Here, $\mathit{SE}(3)$ represents the special Euclidean group of all possible rotations and translations \cite{Szeliski2010} between the two images. The depth estimate $\bm{d}_t$ predicted from $\bm{x}_t$ together with the relative poses $\bm{T}_{t\rightarrow t'}$ can be used to project the image pixels $\bm{x}_{t'}$ to the pixel coordinates of the frame $\bm{x}_{t}$. The loss function is then based on the difference between the projected images $\bm{x}_{t'\rightarrow t}$ and the actual image $\bm{x}_{t}$. As previous approaches \cite{Godard2019, Klingner2021}, we use a mixture of absolute difference and structural similarity (SSIM) difference weighted by a factor $\alpha= 0.85$. The absolute difference $\left\lVert\cdot \right\rVert_1$ is computed over all color channels, while the SSIM difference $SSIM_i\left(\cdot\right)\in \mathbb{I}$ is additionally computed on $3\times3$ patches around the pixel index $i$ of the images:
\begin{align}
	J_t^{\mathrm{depth}} &=  \frac{1}{|\mathcal{I}|}\sum_{i \in\mathcal{I}} \min_{t'\in\mathcal{T}'} \left( \frac{\alpha}{2}\left(1-\mathrm{SSIM}_i\left(\bm{x}_{t}, \bm{x}_{t'\rightarrow t}\right)\right)\right. \notag\\ &+  \left(1-\alpha\right) \frac{1}{C}\left\lVert\bm{x}_{t,i} - \bm{x}_{t'\rightarrow t, i}\right\rVert_1 \Big),
	\label{eq:photometric_loss} 
\end{align}
where we employ the minimum reprojection loss from \cite{Godard2019} to handle occlusions between consecutive frames. Note, that the depth estimate $\bm{d}_{t}$ is optimized implicitly as part of the projection model optimized through (\ref{eq:photometric_loss}). 
\par
Additionally, we follow recent works in imposing a smoothness loss \cite{Godard2017}, allowing sharp depth boundaries only at pixel positions with strong color changes inside the input image. This additional loss is weighted by a factor of $10^{-3}$ as in \cite{Casser2019, Godard2017, Klingner2021}. Also, we apply the auto-masking technique from \cite{Godard2019} to minimize the influence of non-moving dynamic objects or frames. For more details on self-supervised depth estimation, the interested reader is referred to \cite{Godard2019, Zhou2017a}, while \cite{Klingner2021, Klingner2020a} contains more details on the multi-task training with semantic segmentation.
\par

\textbf{Improvement 1: Sequential Training Protocol}:
When training a multi-task network, several strategies can be applied: First, both decoders and the encoder (cf.~Fig.~\ref{fig:multi_task_network}) can be trained at once as in \cite{Klingner2021}, which we dub MT. Alternatively, first one task's decoder is trained and afterwards the second task's decoder is trained, while the encoder weights are kept fixed during the second training stage. If the semantic segmentation network is first and afterwards the depth decoder, we dub this S,D. If we train the tasks in the opposite order, we dub this D,S. In this work we propose the sequential training protocol S,D,MT yielding an improved prediction performance as well as an improved absolute performance compared to the common MT training protocol.
\par
\textbf{Improvement 2: Sharing More Network Layers}:
As shown in Fig.~\ref{fig:multi_task_network}, some layers of the semantic segmentation DNN and the depth estimation DNN can be shared to improve the correlation between both task's outputs. To improve absolute performance, the common approach is to share the encoder and to use task-specific decoders \cite{Klingner2021}. However, for online performance prediction it can be beneficial to even share some decoder layers, if their architecture is identical in the first layers. Thereby, correlation and prediction performance can be improved even further.

\subsection{Online Performance Prediction}

To complete the performance prediction framework, the regression between the depth metric and the segmentation metric is calibrated in an offline regression setup \cite{Klingner2021}. In this work, we then propose to temporally aggregate the statistics from the predicted segmentation performance during online inference to improve prediction performance without compromising segmentation performance.
\par
\begin{figure}
    \centering
    \includegraphics[width=1.0\linewidth]{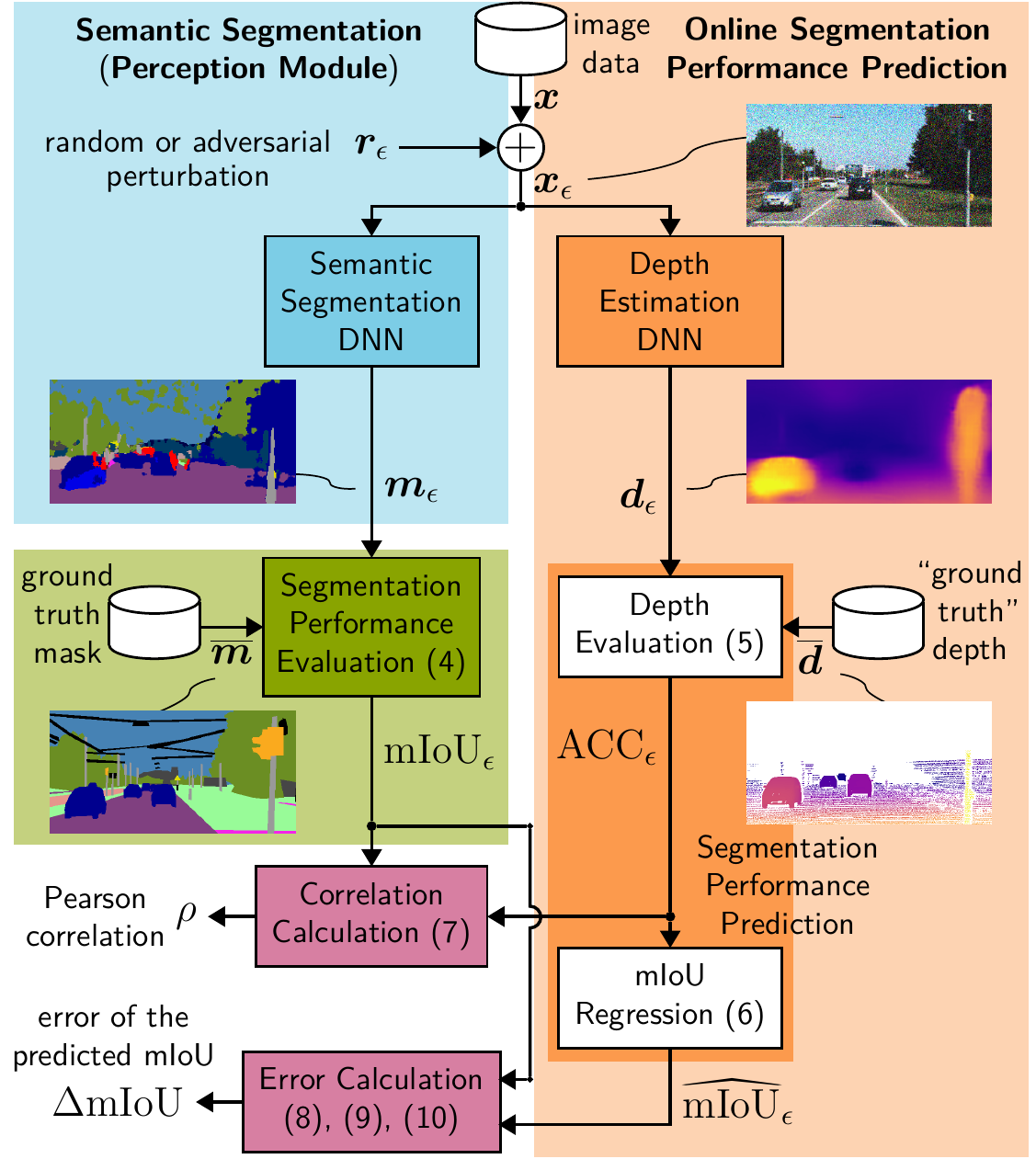}
    \caption{Offline \textbf{test setup} of the \textbf{online performance prediction} for the semantic segmentation task. The input image is artificially perturbed, leading to a performance degradation on both segmentation and depth estimation tasks. This can be used to obtain a regression between the segmentation performance and the depth performance. The depth estimation, the depth evaluation (with LiDAR ``groud truth''), and the regression then make up the online performance prediction during inference, whose quality is reported by correlation and error metrics.}
    \label{fig:offline_regression}
\end{figure}

\textbf{Network Input Perturbations}:
After the network has been trained, we simulate environment changes degrading network performance during inference by using various noise and adversarial attack types as shown in Fig.~\ref{fig:offline_regression}. These are added as perturbation $\bm{r}_{\epsilon}\in \left[-1,1\right]^{H\times W\times C}$ onto the input image $\bm{x}$ as
\begin{equation}
	\bm{x}_\epsilon = \bm{x} + \bm{r}_{\epsilon},
	\label{eq:add_input_perturbation}
\end{equation}
yielding the perturbed input image $\bm{x}_\epsilon$. We then pass the perturbed image through the multi-task network estimating depth $\bm{d}_\epsilon$ and segmentation $\bm{m}_\epsilon$, whose quality decreases dependent on the perturbation strength $\epsilon$. We ensure $\bm{x}_\epsilon\in\mathbb{I}^{H\times W\times C}$ through clipping to the correct range. Moreover, the perturbation strength $\epsilon$ is defined by the signal-to-noise ratio as in \cite{Klingner2020} to make different perturbation types comparable. During evaluation, we apply several $\epsilon \in \mathcal{E}$, with the set of applied perturbation strengths $\mathcal{E}$.
\par
\textbf{Segmentation Evaluation}: 
To implement a regression between the performance of the primary semantic segmentation task and the auxiliary depth estimation task, both tasks need to be evaluated. For semantic segmentation we employ the mean intersection-over-union ($\mathrm{mIoU}$) metric \cite{Everingham2015}, calculated from the number of true positives ($\mathrm{TP}_{\epsilon, s}$), false negatives ($\mathrm{FN}_{\epsilon, s}$), and false positives ($\mathrm{FP}_{\epsilon, s}$) between prediction $\bm{m}_\epsilon$ and ground truth $\overline{\bm{m}}$ for each class $s$ as
\begin{equation}
	\mathrm{mIoU}_\epsilon = \frac{1}{|\mathcal{S}|}\sum_{s\in\mathcal{S}}\frac{\mathrm{TP}_{\epsilon, s}}{\mathrm{TP}_{\epsilon, s} + \mathrm{FP}_{\epsilon, s} + \mathrm{FN}_{\epsilon, s}}.
	\label{eq:miou}
\end{equation}
The clean performance $\mathrm{mIoU} = \mathrm{mIoU}_0$ is given for a perturbation strength $\epsilon = 0$, while the regression is calculated from different $\epsilon \geq 0$. Note that $\mathrm{TP}_{\epsilon, s}$, $\mathrm{FN}_{\epsilon, s}$ and $\mathrm{FP}_{\epsilon, s}$ are usually determined on an entire test set. As in this work performance prediction should, however, be applied on a single-image/few-shot basis, we calculate (\ref{eq:miou}) on a single-image basis and report a test set average then.
\par
\textbf{Depth Evaluation}: The depth estimation task is usally evaluated using seven different metrics (cf. \cite{Eigen2014, Zhou2017a, Godard2017, Casser2019}). For comparability with previous works \cite{Klingner2021}, we pick the accuracy metric $\mathrm{ACC}$ (in literature often referred to as ``$\delta < 1.25$'') for our performance prediction framework. It is calculated on a single-image basis by
\begin{equation}
\mathrm{ACC}_\epsilon = \frac{1}{|\overline{\mathcal{I}}|}\sum_{i\in\overline{\mathcal{I}}} \left[\max\left(\frac{d_{\epsilon, i}}{\overline{d}_{i}}, \frac{\overline{d}_{i}}{d_{\epsilon,i}}\right)\! <\! 1.25\right]_{\mathrm{Iv}}\!,
\label{eq:acc}
\end{equation}
where $d_{\epsilon, i}$ ad $\overline{d}_i$ are the pixel-wise elements of the predicted depth $\bm{d}_\epsilon$ and the ground truth depth $\overline{\bm{d}}$, respectively, $\overline{\mathcal{I}} \subset \mathcal{I}$ is the subset of all pixels for which the ground truth depth is available, and $|\overline{\mathcal{I}}|$ is the number of labeled pixels. The ``ground truth'' depth $\overline{\bm{d}}$ can be thought of as a LiDAR measurement, which is available during inference. Moreover, $\left[\cdot\right]_{\mathrm{Iv}}$ represents the Iverson bracket being $1$ or $0$ if the condition inside the bracket is true or false, respectively. Also note that we determine the median of all image-wise scale factors on the validation set and apply this scale factor on the test set to make up for the global scale ambiguity in self-supervised monocular depth estimation.
\par
\textbf{Offline Regression Setup}:
The aim of the regression is to obtain a predicted performance $\widehat{\mathrm{mIoU}}_{\epsilon}$ from the depth performance $\mathrm{ACC}_{\epsilon}$, which can be used to replace the actual performance $\mathrm{mIoU}_{\epsilon}$ during inference. Assuming a non-linear dependency between both metrics, we apply a polynomial regression of second order 
\begin{equation}
	\widehat{\mathrm{mIoU}}_{\epsilon} = \sum_{k\in \mathcal{K}} \theta_k\cdot \mathrm{ACC}_{\epsilon}^k,
	\label{eq:poly_regression}
\end{equation}
where the parameters $\theta_k$ with $\mathcal{K} = \left\lbrace 0,1,2 \right\rbrace$ have to be optimized on the validation set. To this end, we consider the validation set $\mathcal{X} = \left\lbrace \bm{x}_1,\bm{x}_2,...,\bm{x}_n,... , \bm{x}_{|\mathcal{N}|}\right\rbrace$ with $|\mathcal{N}|$ images from the set of images $\mathcal{N}$, where the actual performance $\mathrm{mIoU}_{n, \epsilon}$ and the predicted performance $\widehat{\mathrm{mIoU}}_{n, \epsilon}$ are calculated for each image $\bm{x}_n$. Accordingly, the regression parameters can be calibrated by minimizing the mean squared error $\sum_{n,\epsilon} (\widehat{\mathrm{mIoU}}_{n, \epsilon}-\mathrm{mIoU}_{n, \epsilon})^2$.
\par
\textbf{Online Inference Setup}: During inference, the depth estimation, the depth evaluation (\ref{eq:acc}) with LiDAR-based ``ground truth'' depth estimates $\overline{\bm{d}}$, and the regression (\ref{eq:poly_regression}) can be used to predict the segmentation performance in online operation as shown in Fig.~\ref{fig:offline_regression} (blue and orange parts only). In this work, we use a test set to simulate online operation and to evaluate our method (+ Fig.~\ref{fig:offline_regression} green and red parts). Thereby, we can also calculate the actual absolute performance by averaging the single-image results from (\ref{eq:miou}) for $\epsilon=0$ over the entire test set.
\par
To evaluate the performance prediction, we use one correlation metric and two error metrics as defined in \cite{Klingner2021}. At first, the Pearson correlation metric is computed as
\begin{equation}
	\rho = \frac{\sum_{n,\epsilon} \left(a_{n,\epsilon} - \mu_a\right) \left(b_{n,\epsilon} - \mu_b\right)}{\sqrt{\sum_{n,\epsilon} \left(a_{n,\epsilon} - \mu_a\right)^2}\sqrt{\sum_{n,\epsilon} \left(b_{n,\epsilon} - \mu_b\right)^2}},\label{eq:pearson_correlation}
\end{equation}
with abbreviations $a_{n,\epsilon}=\mathrm{mIoU}_{n,\epsilon}$ and $b_{n,\epsilon}=\mathrm{ACC}_{n,\epsilon}$, $\rho\in \left[-1,1\right]$, and $\mu_a = \frac{1}{|\mathcal{N}||\mathcal{E}|} \sum_{n,\epsilon} a_{n,\epsilon}$ (likewise for $\mu_b$). A correlation coefficient of $\rho = 1$ indicates perfect positive correlation, while a correlation coefficient of $\rho = -1$ indicates perfect negative correlation. On the other hand, a correlation coefficient of $\rho = 0$ indicates no correlation at all between both entities.
\par
We evaluate the performance prediction by using two error metrics between the predicted performance $\widehat{\mathrm{mIoU}}_{n,\epsilon}$ and the actual performance $\mathrm{mIoU}_{n,\epsilon}$. as shown in Fig.~\ref{fig:offline_regression}. To this end, we first calculate the image-wise differences
\begin{equation}
	\Delta \mathrm{mIoU}_{n,\epsilon} = \widehat{\mathrm{mIoU}}_{n, \epsilon} - \mathrm{mIoU}_{n, \epsilon}.
	\label{eq:differences}
\end{equation}
Afterwards, we calculate the mean absolute error 
\begin{equation}
	\Delta \mathrm{mIoU}^{\mathrm{M}} = \frac{1}{|\mathcal{N}|\cdot |\mathcal{E}|}\sum_{n\in\mathcal{N}}\sum_{\epsilon\in\mathcal{E}} |\Delta \mathrm{mIoU}_{n,\epsilon}|,
	\label{eq:abs_error_mae}
\end{equation}
as well as the root mean squared error
\begin{equation}
	\Delta \mathrm{mIoU}^{\mathrm{R}} = \sqrt{\frac{1}{|\mathcal{N}|\cdot |\mathcal{E}|}\sum_{n\in\mathcal{N}}\sum_{\epsilon\in\mathcal{E}} \left(\Delta \mathrm{mIoU}_{n,\epsilon}\right)^2}.
	\label{eq:abs_error_rmse}
\end{equation}
The mean absolute error gives a very intuitive idea of the deviation between actual and predicted performance, while the root mean squared error is more sensitive to outliers.
\par
\textbf{Improvement 3: Aggregation of Statistics over Time}:
While it is of course interesting to obtain performance predictions on a single-image basis, in this work we aim at improving performance by aggregation of statistics over a longer window of $\Delta N$ frames. To acquire these $\Delta N$ frames with indices $\nu\in\mathcal{N}_n$, every $K$-th frame is taken with frame $n$ being in the center: $\mathcal{N}_n = \left\lbrace n\!-\!\frac{\Delta N -1}{2}\!\cdot\! K, ..., n\!-\!K, n, n\!+\!K, n\!+\!K\!\cdot\!\frac{\Delta N - 1}{2} \right\rbrace$. However, this comes at the cost of an algorithmic latency of $K\cdot\frac{\Delta N - 1}{2}$ frames, or $\tilde{T} = \frac{\Delta N - 1}{2 f}\left[ s \right]$ with $f\left[\frac{1}{s}\right]$ being the frame rate of the deployed camera. This can be particularly tolerated in slowly changing environments such as, \eg, day/night shifts. In that case, one can use (predicted) quality averages over these time instants
\begin{align}
  \mathrm{mIoU}^{(\Delta N)}_{n, \epsilon} &= \frac{1}{\Delta N}\cdot \sum_{\nu\in\mathcal{N}_n} \mathrm{mIoU}_{\nu, \epsilon} \label{eq:averaging_gt}  \\
  \widehat{\mathrm{mIoU}}^{(\Delta N)}_{n, \epsilon} &= \frac{1}{\Delta N}\cdot \sum_{\nu\in\mathcal{N}_n} \widehat{\mathrm{mIoU}}_{\nu, \epsilon} \label{eq:averaging_pred}
\end{align}
for the actual and the predicted performance, respectively. Accordingly, the quantity $\widehat{\mathrm{mIoU}}^{(\Delta N)}_{n, \epsilon}$ does represent the mean (not the single-image) predicted performance for $\Delta N$ frames $\bm{x}_{\nu}, \nu\in\mathcal{N}_n$. As we intend to apply this method during offline testing on labelled image data (not necessarily videos), we randomly sample $\frac{\Delta N -1}{2}$ ``preceding'' and $\frac{\Delta N -1}{2}$ ``successive'' images indexed by $\mathcal{N}_n$ from the whole test set. However, as these images are uncorrelated, we need to show which time $\frac{K_{\mathrm{min}}}{f}$ between sampled video frames $n, n\!+\!K_{\mathrm{min}},...$ is necessary for this assumption to hold. The error of the prediction is then given by the difference
\begin{equation}
	\Delta \mathrm{mIoU}_{n,\epsilon} = \widehat{\mathrm{mIoU}}^{(\Delta N)}_{n, \epsilon} - \mathrm{mIoU}^{(\Delta N)}_{n, \epsilon},
	\label{eq:differences2}
\end{equation}
and subsequent application of (\ref{eq:abs_error_mae}) and (\ref{eq:abs_error_rmse}). By this technique, one can improve the performance prediction at the cost of a certain algorithmic latency $\tilde{T}$ during inference.

\section{Experimental Setup}
\label{sec:implementation_details}

Describing our experimental setup, we outline our employed datasets, network architecture, training procedure, and evaluation metrics. All models are implemented in \texttt{PyTorch} \cite{Paszke2019} and have been trained and evaluated on an \texttt{NVIDIA Tesla V100} graphics card.
\par
\textbf{Employed Datasets}:
During our experiments we have three stages: the training stage, the regression stage and the testing stage, which all require different datasets. For the training stage, we use the KITTI dataset \cite{Geiger2013} with the KITTI split $\tilde{\mathcal{X}}^{\mathrm{KIT}}_{\mathrm{train}}$ defined by \cite{Godard2017} to train the depth estimation part of our method. Note, that due to the requirement of a preceding and a succeeding frame for our self-supervised training scheme we can use only 28,937 of the originally defined 29,000 images. Meanwhile, we use the 2,975 images of the training split $\mathcal{X}^{\mathrm{CS}}_{\mathrm{train}}$ from the Cityscapes dataset \cite{Cordts2016} to train the segmentation in a supervised fashion.
\par
To evaluate our online performance prediction technique we need two data subsets with available labels for both depth and segmentation. One subset is used during the regression stage and the other one during the test stage. To achieve this, we split the 200 images of the KITTI Stereo 2015 training set \cite{Menze2015} having the necessary labels for both tasks into two subsets. We use 50 images ($\mathcal{X}^{\mathrm{KIT}}_{\mathrm{val}}$) to determine the regression parameters and 150 images ($\mathcal{X}^{\mathrm{KIT}}_{\mathrm{test}}$) to test and evaluate our method.
\par
\textbf{Network Architecture}:
Our semantic segmentation DNN architecture is based on the encoder-decoder architecture with skip connections from \cite{Godard2019}. We follow \cite{Klingner2020a} in using the same architecture for the depth estimation DNN. The only difference is the output layer, where the depth decoder has a sigmoid output $\sigma_{n,i}$, where the final depth map is created by $\frac{1}{a \sigma_{n,i} + b}$ with $a$ and $b$ constraining the pixel-wise depth estimates in the range $\left[0.1, 100\right]$. The segmentation decoder on the other hand uses a softmax activation function on the $|\mathcal{S}|$ feature maps such that the logits are converted to class probabilities. Using the same architecture for both tasks enables investigations on how many layers should be shared for online performance prediction.
\par
\textbf{Training Details}:
We train both tasks simultaneously using mixed batches with 6 images from both the depth and the segmentation dataset, i.e., 12 images in total per minibatch. These are all passed through the multi-task network, while the losses are only calculated on the task-specific images. Additionally, for depth estimation, we use the full-resolution multi-scale loss from \cite{Godard2019} and apply gradient scaling with factors of $0.9$ and $0.1$ for depth and segmentation, respectively, before the gradients reach the shared network parts. For optimization we apply the Adam optimizer \cite{Kingma2015} for $40$ epochs with a learning rate of $10^{-4}$, which is reduced to $10^{-5}$ after $30$ epochs. For more training details the interested reader is referred to \cite{Klingner2021, Klingner2020a}.
\par
\textbf{Evaluation Details}:
During evaluation we apply four different perturbation types, namely Gaussian noise, salt and pepper noise \cite{Gonzalez2008}, FGSM attacks \cite{Goodfellow2015}, and PGD attacks \cite{Madry2018}. As proposed by \cite{Klingner2021}, we use a mix of them to calibrate the regression model and the mean over all four perturbation types to evaluate the performance prediction. We apply perturbation strengths of $\epsilon \in\mathcal{E} = \left\lbrace 0.25, 0.5, 1, 2, 4, 8, 12, 16, 20, 24, 28, 32 \right\rbrace\cdot \frac{1}{255}$ to cover the whole performance range of the segmentation model.

\section{Experimental Evaluation}
\label{sec:experiments}

In this section, we will first evaluate and analyze the details of our regression used as part of the performance prediction framework. Afterwards, we outline how the performance prediction error can be reduced through a sequential training protocol, sharing of more layers, and (predicted) performance aggregation over a number of frames.

\subsection{Regression Calibration}

\begin{figure}
    \centering
    \resizebox{\linewidth}{!}{\input{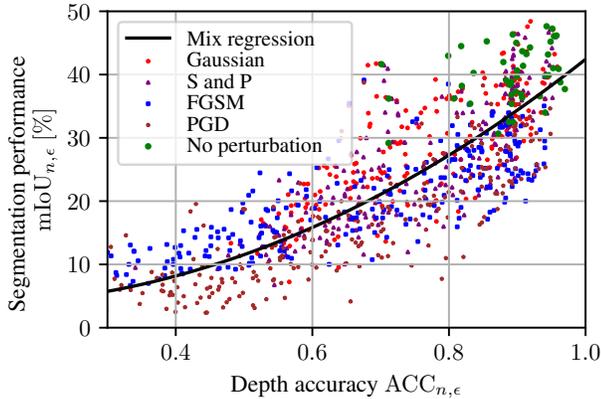}}
    \caption{\textbf{Regression analysis}: We show a scatter plot of the image-wise calculated metrics $\mathrm{ACC}_{n,\epsilon}$ (\ref{eq:acc}) and $\mathrm{mIoU}_{n,\epsilon}$ (\ref{eq:miou}) on 20 images of the KITTI validation set $\mathcal{X}^{\mathrm{KIT}}_{\mathrm{val}}$. The results without any perturbation are marked as green circles. Also, we show the mix regression result (black curve), which is obtained using four different perturbation types and various perturbation strengths $\epsilon$.}
    \label{fig:regression}
    \vspace{-0.2cm}
\end{figure}

Analyzing our performance prediction method in Fig.~\ref{fig:regression}, we show the dependency of the semantic segmentation metric $\mathrm{mIoU}_{n, \epsilon}$ from the depth metric $\mathrm{ACC}_{n, \epsilon}$, the value pairs being obtained on $20$ images $\bm{x}_n$ perturbed with perturbations of strength $\epsilon\in \mathcal{E}$. For this experiment we use a model trained in a multi-task fashion as proposed by \cite{Klingner2021}. We can observe that the mix regression model (black curve), obtained on the validation set using results from all four perturbation types, is able to follow the trend of the data points to some extent. Surprisingly, we can also observe that the depth and segmentation metrics on clean images ($\epsilon=0$, cf. green circles in Fig.~\ref{fig:regression}) are not highly correlated as can be seen from their widely distributed positions in Fig.~\ref{fig:regression}. Our conclusion is therefore that our method does indeed not exploit an intrinsic correlation on clean images between two task's metrics. Our method rather relies on the correlation of the performance \textit{degradation} in both tasks' outputs. As can be seen from the coherence in which the mix regression model captures the single-image results (cf. the regression in Fig.~\ref{fig:regression}), this correlation is similar for different images and for different perturbation types and strengths.

\subsection{Influence of Different Training Protocols}

Now, we investigate which training protocol is best suited for our performance prediction method with results shown in Tab.~\ref{tab:training_techniques}. The advantage of only changing the training protocol is that no computational overhead is induced during inference and the network structure and complexity remain unchanged. When comparing the standard multi-task (MT) model to the S,D model, where first the segmentation network is trained and afterwards only the depth decoder, we observe a better prediction performance of the latter ($\Delta\mathrm{mIoU}^{\mathrm{M}} = 4.45\%$ compared to $\Delta\mathrm{mIoU}^{\mathrm{M}} = 6.01\%$) at the cost of a lower absolute performance ($\mathrm{mIoU} = 34.97\%$ compared to $\mathrm{mIoU} = 37.51\%$). By running S,D with subsequent MT in the S,D,MT training protocol, we can combine the strengths of both training schemes, i.e., we obtain a quite low performance prediction error $\Delta\mathrm{mIoU}^{\mathrm{M}} = 5.33\%$, while having an even higher absolute performance than the MT model with $\mathrm{mIoU} = 39.43\%$. Please note that the best numbers of the depth performance $\mathrm{ACC}$ in Tabs.~\ref{tab:training_techniques} and \ref{tab:shared_layers} are not written in boldface as we do not aim at optimizing this secondary task.
\par
\begin{table}[t]
\caption{\textbf{Different training protocols}: Results on $\mathcal{X}^{\mathrm{KIT}}_{\mathrm{test}}$ for multi-task training (MT), single-task depth training (D), and single-task segmentation training (S). If a second single-task training is applied (S after D or D after S), only the second decoder is trained. mIoU values in $\%$; best results in \textbf{boldface}.}
\label{tab:training_techniques}
\centering
\small
\setlength{\tabcolsep}{5pt}
\begin{tabular}{c|cc|ccc}
	Training & \multicolumn{2}{c|}{Performance} & \multicolumn{3}{c}{Prediction}\stz\\ 
	protocol & $\mathrm{ACC}$ & $\mathrm{mIoU}$ & $\rho$ & $\Delta \mathrm{mIoU}^{\mathrm{M}}$ & $\Delta \mathrm{mIoU}^{\mathrm{R}}$\stz\\
	\hline
	MT \cite{Klingner2021} & $0.85$	& $37.51$	& $0.77$	& $6.01$	& $7.70$\stz\\
	S,D \cite{Klingner2021} & $0.86$	& $34.97$	& $\textbf{0.86}$	& $4.45$	& $\textbf{6.16}$\stz\\
	S,MT & $0.83$	& $\textbf{39.87}$	& $0.81$	& $6.08$	& $7.68$\stz\\
	S,D,MT & $0.86$	& $39.43$	& $0.84$	& $5.33$	& $7.01$\stz\\
	D,S \cite{Klingner2021} & $0.84$	& $33.34$	& $0.77$	& $5.49$	& $7.34$\stz\\
	D,MT & $0.82$	& $39.81$	& $0.78$	& $6.14$	& $8.00$\stz\\
	D,S,MT & $0.82$	& $36.57$	& $0.77$	& $5.64$	& $7.51$\stz\\
\end{tabular}
\end{table}

Still, the separate training of the depth decoder seems to be important, as the S,MT model has a similar absolute performance ($\mathrm{mIoU} = 39.87\%$ compared to $\mathrm{mIoU} = 39.43\%$) as the S,D,MT model, but a higher performance prediction error ($\Delta\mathrm{mIoU}^{\mathrm{M}} = 6.08\%$ compared to $\Delta\mathrm{mIoU}^{\mathrm{M}} = 5.33\%$). Also, training the depth network in a standalone fashion in the beginning (D,S model, D,MT model, and D,S,MT model in Tab.~\ref{tab:training_techniques}) offers no clear advantages over the MT model. Concluding, we show that \textit{with the S,D,MT model we improve on the work of \cite{Klingner2021} (proposing the MT model) in terms of both performance prediction ($\Delta \mathrm{mIoU}$ metrics) and absolute performance ($\mathrm{mIoU}$) without computational overhead during inference}. 

\subsection{Influence of Sharing Network Parts}

\begin{table}[t]
\caption{\textbf{Sharing network parts}: Results on $\mathcal{X}^{\mathrm{KIT}}_{\mathrm{test}}$ when different \textbf{network parts} are \textbf{shared} during training. We report results for completely separately trained networks, a shared encoder, as well as additionally shared decoder layers. All multi-task networks are trained according to the \textbf{S,D,MT} training protocol proposed in Tab.~\ref{tab:training_techniques}. mIoU values in $\%$; best results in \textbf{boldface}.}
\label{tab:shared_layers}
\centering
\small
\setlength{\tabcolsep}{3.8pt}
\begin{tabular}{l|cc|ccc}
	\quad\; Shared & \multicolumn{2}{c|}{Performance} & \multicolumn{3}{c}{Prediction}\stz\\ 
	network parts & $\mathrm{ACC}$ & $\mathrm{mIoU}$ & $\rho$ & $\Delta \mathrm{mIoU}^{\mathrm{M}}$ & $\Delta \mathrm{mIoU}^{\mathrm{R}}$\stz\\
	\hline
	None \cite{Klingner2021} & $0.84$	& $34.97$	& $0.67$	& $7.55$	& $10.10$\stz\\
	+ Encoder & $0.86$	& $39.43$	& $0.84$	& $5.33$	& $7.01$\stz\\
	+ Dec. layer 1 & $0.86$	& $\textbf{40.79}$	& $0.82$	& $5.87$	& $7.71$ \stz\\
	+ Dec. layer 2 & $0.86$	& $38.71$	& $0.82$	& $5.67$	& $7.30$ \stz\\
    + Dec. layer 3 & $0.86$	& $35.80$	& $0.83$	& $4.96$	& $6.48$ \stz\\
    + Dec. layer 4 & $0.86$	& $33.94$	& $\textbf{0.88}$	& $\textbf{4.07}$	& $\textbf{5.37}$ \stz\\
\end{tabular}
\end{table}

Another way to improve the performance prediction is to increase the correlation between both task's outputs by sharing more network layers. The results in Tab.~\ref{tab:shared_layers}, where all models with shared network parts are trained according to the S,D,MT training protocol, clearly show that it is beneficial to share the encoder layers, as absolute performance improves from $\mathrm{mIoU} = 34.97\%$ to $\mathrm{mIoU} = 39.43\%$ and at the same time the prediction error reduces from $\Delta\mathrm{mIoU}^{\mathrm{M}} = 7.55\%$ to $\Delta\mathrm{mIoU}^{\mathrm{M}} = 5.33\%$. Sharing one additional decoder layer improves the absolute performance slightly but surprisingly not the performance prediction error. Only when sharing three or four decoder layers, we can reduce the performance prediction error significantly down to $\Delta\mathrm{mIoU}^{\mathrm{M}} = 4.07\%$, however, at the cost of a decreased absolute performance of $\mathrm{mIoU} = 33.94\%$. This could be because the decoders cannot learn optimal task-specific features in the reduced remaining number of task-specific layers. In conclusion, \textit{sharing only the encoder already offers a good trade-off between absolute performance and performance prediction quality}, which is why we will stick to this model for our further analysis. If one aims at a significantly improved performance prediction or a more efficient performance prediction with less computational overhead, sharing more decoder layers is also a good option.

\subsection{Influence of Aggregation of Statistics}

\begin{figure}
    \centering
    \resizebox{\linewidth}{!}{\input{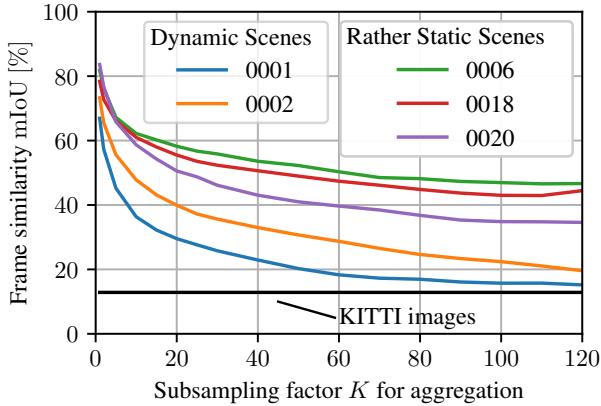}}
    \caption{\textbf{Temporal correlation analysis}: Overview on the average $\mathrm{mIoU}$ between pairs of samples stemming from the KITTI dataset or from videos of the vKITTI dataset subsampled to every $K$-th frame. We observe that subsampling dynamic scenes by a factor of $K$ in the range $K=100...200$ yields similar $\mathrm{mIoU}$ for an image pair as any KITTI image pair.}
    \vspace{-0.2cm}
    \label{fig:frame_averaging}
\end{figure}

Here, we will evaluate our simple aggregation method described by (\ref{eq:averaging_gt}), (\ref{eq:averaging_pred}), and (\ref{eq:differences2}), where the mean performance over $\Delta N$ frames is predicted. As labeled real-world video sequences are hard to obtain for semantic segmentation, we calculate the $\mathrm{mIoU}$ between any pair of images of the KITTI dataset as shown by the black line in Fig.~\ref{fig:frame_averaging}. As these frames are rather uncorrelated, we use the video sequences of the vKITTI dataset \cite{Gaidon2016} to check how large we have to choose $K$ so that in a real world image sequence the same dissimilarity is obtained as if we used a random number of $\Delta N \!-\! 1$ extra images from our KITTI data for aggregation (\ref{eq:averaging_gt}), (\ref{eq:averaging_pred}). We can see that on the sequences $0006$, $0018$, and $0020$ the approximation does not hold as these videos either show highway scenes with few variations or static scenes of a barely moving vehicle. On the other hand, for the more dynamic videos $0001$ and $0002$, the $\mathrm{mIoU}$ approaches the one present in the KITTI 2015 dataset (where segmentation ground truth is available) for a subsampling factor of $K=100...200$. For this distance the evaluation on the KITTI 2015 dataset should be approximating a real video (for dynamically changing scenes).
\par

\begin{table}[t]
\caption{\textbf{Aggregation of statistics}: Results on $\mathcal{X}^{\mathrm{KIT}}_{\mathrm{test}}$ when the actual performance $\mathrm{mIoU}$ and the predicted performance $\widehat{\mathrm{mIoU}}$ are averaged over several frames before the calculation of the difference in (\ref{eq:differences}). We report results for different numbers of $\Delta N$ frames used for aggregation (first column). mIoU values in $\%$.}
\label{tab:frame_averaging}
\centering
\small
\setlength{\tabcolsep}{1.8pt}
\begin{tabular}{c|c|cc|ccc}
	\multirow{2}{*}{$\Delta N$} & Decision & \multicolumn{2}{c|}{Performance} & \multicolumn{3}{c}{Prediction}\stz\\ 
	 & latency $\tilde{T}\left[ s \right]$ & $\mathrm{ACC}$ & $\mathrm{mIoU}$ & $\rho$ & $\Delta \mathrm{mIoU}^{\mathrm{M}}$ & $\Delta \mathrm{mIoU}^{\mathrm{R}}$\stz\\
	\hline
	1 & $0$ & $0.86$	& $39.43$	& $0.84$	& $5.33$	& $7.01$ \stz\\
	3 & $10$ & $0.86$	& $39.43$	& $0.92$	& $3.88$	& $4.91$ \stz\\
	5 & $20$ & $0.86$	& $39.43$	& $0.94$	& $3.30$	& $4.17$ \stz\\
    11 & $50$ & $0.86$	& $39.43$	& $0.95$	& $3.02$	& $3.77$ \stz\\
    21 & $100$ & $0.86$	& $39.43$	& $0.96$	& $2.81$	& $3.51$ \stz\\
    51 & $250$ & $0.86$	& $39.43$	& $0.96$	& $2.74$	& $3.42$ \stz\\
    101 & $500$ & $0.86$	& $39.43$	& $\textbf{0.97}$	& $\textbf{2.68}$	& $\textbf{3.34}$ \stz\\
\end{tabular}
\end{table}

The monitored time interval is given by $\Delta T$ = $\frac{\Delta N \cdot K}{f}$, and the decision latency in Tab.~\ref{tab:frame_averaging} is $\tilde{T}=\frac{\Delta N -1 }{2}\cdot \frac{K}{f}$. The frame rate of the KITTI and vKITTI datasets is constant with $f=\SI{10}{s^{-1}}$ and we assume $K=100$ frames distance. We can see that the mIoU prediction error decreases significantly already when considering only $\Delta N = 3$ frames, as the effect of outliers in both the predicted and the actual performance is reduced. and even more for a higher number of frames. A good trade-off is given by using $\Delta N= 11$ frames, where we get a significantly reduced prediction error of $\Delta\mathrm{mIoU}^{\mathrm{M}} = 3.02\%$ compared to $\Delta\mathrm{mIoU}^{\mathrm{M}} = 5.33\%$, while the absolute performance does obviously not change. This comes with a decision latency of $\tilde{T}=\SI{50}{s}$, which can be tolerated for slowly changing perturbations such as day/night changes, while for more rapid environment changes the single-image prediction method without aggregation should be preferred.

\section{Conclusion}
\label{sec:conclusion}

In this work we introduce a concept for online performance prediction of semantic segmentation DNNs based on the monocular estimation of depth, its subsequent evaluation using, \eg, LiDAR output, and finally a regression to predict the segmentation performance. In particular, we show how the prediction error can be reduced by sequential training of both tasks, sharing encoder and decoder layers inside a multi-task network, and temporal aggregation of statistics. Thereby, we can reduce the online $\mathrm{mIoU}$ prediction error from $6.01\%$ down to $5.33\%$ (\ie, $11\%$ relative improvement), while even improving the absolute $\mathrm{mIoU}$ performance from $37.51\%$ to $39.43\%$. At the cost of some decision latency, the $\mathrm{mIoU}$ error can be further reduced to $3.02\%$ absolute or even below. Our online performance prediction concept is straightforward extendable to other perception tasks such as object detection or optical flow and provides valuable information for high-level decision systems inside safety critical applications such as autonomous driving or virtual reality.

\clearpage
{\small
\bibliographystyle{ieee_fullname}
\bibliography{bib/ifn_spaml_bibliography}
}

\end{document}